\documentclass[runningheads]{llncs}
\usepackage{graphicx}

\usepackage{tikz}
\usepackage{comment}
\usepackage{amsmath,amssymb} 
\usepackage{color}
\usepackage{xcolor}
\usepackage{colortbl}
\usepackage{array}
\usepackage{multirow}
\usepackage{booktabs}
\usepackage[colorlinks,
            linkcolor=red,      
            anchorcolor=green,  
            citecolor=blue, 
            ]{hyperref}

\usepackage[accsupp]{axessibility}  

\usepackage[width=122mm,left=12mm,paperwidth=146mm,height=193mm,top=12mm,paperheight=217mm]{geometry}

\definecolor{grayDark}{gray}{0.93}
\definecolor{grayLightDark}{gray}{0.96}
\definecolor{grayLight}{gray}{0.99}

\begin{document}
\pagestyle{headings}
\mainmatter
\def\ECCVSubNumber{6684}  

\title{PoseGU: 3D Human Pose Estimation with Novel Human Pose Generator and Unbiased Learning} 


\titlerunning{PoseGU}
%
\author{Shannan GUAN\inst{1} \and
Haiyan LU\inst{1} \and
Linchao ZHU\inst{1} \and
Gengfa FANG\inst{2}}
\authorrunning{GUAN et al.}
%
\institute{Australia Artificial Intelligence Institute \and
School of Electrical and Data Engineering \\
University of Technology Sydney, Australia, AU}
\maketitle

\begin{abstract}
3D pose estimation has recently gained substantial interests in computer vision domain.
Existing 3D pose estimation methods have a strong reliance on large size well-annotated 3D pose datasets, and they suffer poor model generalization on unseen poses due to limited diversity of 3D poses in training sets. 
In this work, we propose \textit{PoseGU}, a novel human pose generator that generates diverse poses with access only to a small size of seed samples, while equipping the Counterfactual Risk Minimization to pursue an unbiased evaluation objective. 
Extensive experiments demonstrate \textit{PoseGU} outforms almost all the state-of-the-art 3D human pose methods under consideration over three popular benchmark datasets.
Empirical analysis also proves \textit{PoseGU} generates 3D poses with improved data diversity and better generalization ability. 

\end{abstract}

\section{Introduction}\label{intro}
3D human pose estimation aims to predict the $x$-$y$-$z$ positions of key joints on a human body in a camera coordinate system by giving a 2D pose as input. 
Conventional way to tackle the 3D pose estimation task is to use a learning-based method ~\cite{stgcn,posemachine,semigcn,mlp} along with  well-annotated 3D poses as training dataset, e.g., Human3.6M~\cite{H36M}, and MPI-INF-3DHP~\cite{3DHP}. 
Despite their success on laboratorial environment,
well-annotated datasets are hard to acquire since pose annotations depend on high-reliable motion capture systems which cannot be easily deployed in real-world environments. 
Besides, pose estimators trained with ground truth datasets are hard to generalize to unknown poses (i.e., poses in the wilds) due to the limited diversity of training dataset, body sizes, and camera views. 

Existing efforts have mainly devoted to overcome these obstacles from roughly two perspectives: 
1) Adopt weakly/self supervised mechanism to reduce the reliance on well-annotated 3D pose data ~\cite{adver1,adver2,HMR,unsuper1,unsuper2}. Due to inaccurate supervisions~\cite{adver1}, these methods lead to a sub-optimal performance on estimating 3D poses compared with fully-supervised learning methods. 
2) Resort to data augmentation to contrast the observational ground truth poses with the generated ones~\cite{aug1,aug2,aug3,poseaug},
which still cannot escape from the heavy reliance on large size well-annotated 3D pose data. 

\begin{figure}[h]
    \centering
    \includegraphics[width=1\linewidth]{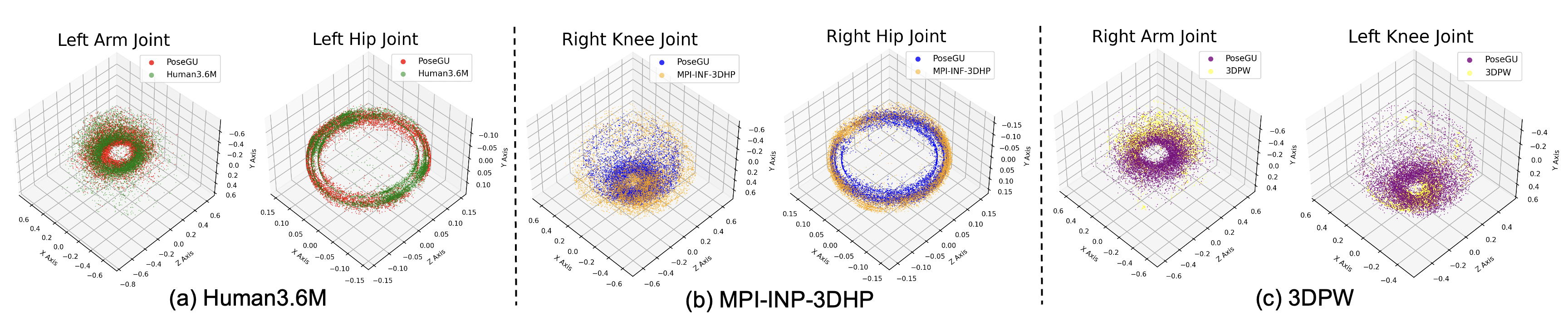}
    \caption{Visualization of data distribution between ground truth source dataset and the generated dataset based on its sampled templates. Each scatter shows the 3D position of a single joint in camera coordinate system.}
    \label{fig:datadistribute}
\end{figure}

To alleviate the aforementioned obstacles, this study proposes a novel 3D pose generator that generates high-quality 3D pose data with access only to minimum seed samples.
The 3D pose generator produces a training dataset composed of the generated 3D poses, while the 3D poses maintain their high data diversity. 
However, it is not trivial to directly leverage the generated training dataset for 3D pose estimation, since the pose distribution of the generated dataset is different from the ground truth dataset and suffers from the well-known ``popularity bias'' issue~\cite{Unbiasedtraining3}. 
An empirical study conducted on our generated dataset and the three ground truth datasets can validate this claim by Figure~\ref{fig:datadistribute}: pose distribution of the generated dataset holds different data distribution compared to the distributions of the three ground truth datasets. 
This is because poses in the ground truth dataset display a skewed popularity distribution across different action categories, leading to a well-known ``popularity bias'' issue~\cite{Unbiasedtraining3}. 
For example, standing poses exist in most of actions, but sitting poses only exist in site-related actions. 
Since the generated dataset from our 3D pose generator evenly takes seed sample from each action category as the input, the generated 3D poses can also suffer from the trap of popularity bias.
In such a case, training pose estimators based on the generated (i.e., biased) dataset cannot reach an unbiased estimation, leading to a sub-optimal performance of estimating 3D poses. 
To address this problem, we further take advantages of Counterfactual Risk Minimization (CRM) to correct the distribution mismatch between ground truth and generated 3D poses datasets to achieve unbiased learning of estimator, such that to guide a better performance of 3D pose estimation.

Our novel contributions are summarized as follows: 
1) Our proposed \textit{PoseGU} has less reliance on ground truth 3D poses and achieves even better performance than conventional methods that directly use a complete ground truth dataset. 
2) To the best of our knowledge, we are the first to develop a 3D human pose generator that can generate 3D pose dataset with high level diversity to train 3D pose estimators with access only to minimum seed samples from ground truth dataset. 
3) We are the first to introduce Counterfactual Risk Minimization (CRM) to achieve unbiased learning in 3D pose estimation task.

\section{Related Work}\label{rework} 
\subsubsection{3D human Pose Estimation} 
Recent 3D human pose estimation works mostly on developing various pose estimators~\cite{mlp,semigcn,videopose,stgcn}. 
These methods are usually trained on well-annotated training data while their performance crucially depends on the availability of the annotated poses that are identical for pose estimation~\cite{survey}. 
Moreover, existing datasets are limited in their scope and variability of the contained poses, making them hard for the trained pose estimators to generalize to unseen poses with different properties~\cite{post1,aug1,poseaug}.
Therefore, some researches tried to improve models’ generalization ability by exploring external information, e.g. some methods utilize unlabeled 2D pose data as a training dataset to train the estimator through adversarial training method~\cite{adver1,adver2}, and use kinematics to regulate or post-processing estimated poses~\cite{post1,post2,post3}. 
More recent works try to increase the pose diversity by utilizing data augmentation method, i.e., Li et al.,~\cite{aug1} directly augment 3D pose data in ground truth dataset by randomly recombining partial skeleton among different 3D poses and perturbing their joint angles, Gong et al.,~\cite{poseaug} expand data diversity of 3D pose data by exploring a learnable method to augment data in ground truth dataset. 
However, these works still highly rely on a large amount of real-world pose data. Unlike all these methods, our proposed 3D pose generator generates infinite diverse 3D pose data by only taking minimum seed samples from ground truth dataset.
To the best of our knowledge no attempt has been made to establish a high-quality pose generator aiming to generate training poses that cover a wide pallet of challenges for human pose estimation.

\subsubsection{Unbiased Learning} 
Previous works~\cite{Unbiasedtraining1} have proved that available datasets have a strong build-in bias, which leads to sub-optimal performance of the downstream models.
The computer vision community has long dealt with the dataset bias issue~\cite{Unbiasedtraining1}. 
However, little effort has been devoted to the dataset bias in 3D pose estimation task. 
Existing 3D pose estimation methods alleviate the dataset bias (i.e., unbiased) mainly focusing on: 
1) performing data re-sampling strategies. 
For example, Yamada et al.,~\cite{Unbiasedtraining3} re-weight training samples by assigning each sample with their propensity scores to achieve the unbiased learning.  
and 2) designing training strategies or unbiased loss functions ~\cite{Unbiasedtraining3,Unbiasedtraining2}. 
Our work combines the idea of re-weight training samples and elaborately design loss functions to achieve unbiased learning.
More specifically, we resort to the Counterfactual Risk Minimization (CRM)~\cite{crmdebias1,crmdebias2} in our objective to directly minimize an empirical risk estimated from the original training dataset as it came from the true risk of the generated dataset, therefore fundamentally removing the distribution mismatch between the original and the generated dataset. 
The CRM has been wildly adopted in computer vision community for unbiased learning~\cite{crmcv1,crmcv2,Unbiasedtraining4}, which is however not been well studied in 3D pose estimation tasks. 

\section{Preliminary}
We introduce the 3D pose estimation problem and a new definition of 3D pose generation in this section. 
\subsection{3D Pose Estimation Problem}
Let $\boldsymbol{x}$ be a $J$-element vector where $J$ is the number of joints in a human body. 
Given $\boldsymbol{x} \in \mathbb{R}^{2 \times J}$ which is 2D coordinates of $J$ joints of human body and its corresponding 3D joint positions $\mathbf{X} \in \mathbb{R}^{3 \times J}$ as derived from a camera coordinate system. 
The objective of 3D pose estimation is to obtain a 3D pose estimator $\mathcal{P}: \boldsymbol{x} \mapsto \mathbf{X}$ that estimates 3D pose $\mathbf{X}$ from the input 2D pose $\boldsymbol{x}$.
Generally, the $\mathcal{P}$ with parameter $\theta$ is trained with 2D-3D pose pair of $\mathcal{X} =\{\boldsymbol{x}, \mathbf{X}\}$ by solving:

\begin{equation}\label{obj1}
\min _{\theta} \mathcal{L}_{\mathcal{P}}\left(\mathcal{P}_{\theta}, \mathcal{X}\right)=\mathcal{L}_{\mathcal{P}}\left(\mathcal{P}_{\theta}(\boldsymbol{x}), \mathbf{X}\right)
\end{equation}
where the loss function $\mathcal{L}_{\mathcal{P}}$ is typically defined as mean square errors (MSE) between estimated and ground truth 3D poses. 

The conventional way to tackle the 3D pose estimation problem, which takes the pose pair $\mathcal{X}$ retained in the given $K$ poses ground truth datasets $\mathcal{D}=\left\{\mathcal{X}_{k} \mid \forall k \in\{1, \cdots, K\}\right\}$, however suffers from the well-known ``poor generalization'' trap when applied to a new dataset. 
The main reason is the limited diversity of the training datasets, in which variations in body sizes, camera intrinsic and extrinsic parameters are not diverse enough to accommodate cross-dataset generalization. 
Next, we give the formal definition of our novel 3D pose generation as follows.

\subsection{3D Pose Generation}
In our work, we give the inherent factors of human pose, i.e. rotation angles and bone lengths, we denote bone length vector of $M$ bones as $\boldsymbol{l} \in \mathbb{R}^{1\times M}$, and represent rotation angles as rotation angle matrix $\mathbf{A} \in \mathbb{R}^{3 \times M}$ to embed angle information under $x$-$y$-$z$ axis using camera coordinate system. 
With our proposed forward kinematics based 3D pose generator $\mathcal{G}: \mathbf{A},\boldsymbol{l} \mapsto \widetilde{\mathcal{X}}$, infinite number of various 3D poses $\widetilde{\mathbf{X}}$ can be generated as training samples to train the pose estimator $\mathcal{P}$. 

Our generated poses are model-agnostic and can be easily applied to any downstream 3D pose estimators (e.g., video pose~\cite{videopose}, semi-gcn~\cite{semigcn}). Moreover, our generated $\widetilde{\mathbf{X}}$ relies purely on the inherent factors (i.e., angles and bone length) of human poses without depending on any coordinate in the existing ground truth datasets, the trained pose estimator $\mathcal{P}$ under our $\widetilde{\mathbf{X}}$ can escape from the curse of the ``poor generalization''. With the generated $\widetilde{\mathbf{X}}$ and the corresponding relation pose pair $\widetilde{\mathcal{X}} =\{\tilde{\boldsymbol{x}}, \widetilde{\mathbf{X}}\}$, the objective function for solving $\mathcal{P}$ can be revised as:

\begin{equation}\label{obj2}
\min _{\theta} \mathcal{L}_{\mathcal{P}}\left(\mathcal{P}_{\theta}, \widetilde{\mathcal{X}}\right)=\mathcal{L}_{\mathcal{P}}\left(\mathcal{P}_{\theta}(\tilde{\boldsymbol{x}}), \widetilde{\mathbf{X}}\right)
\end{equation}

\section{Methodology}\label{method} 

\begin{figure}[h]
    \centering
    \includegraphics[width=1\linewidth]{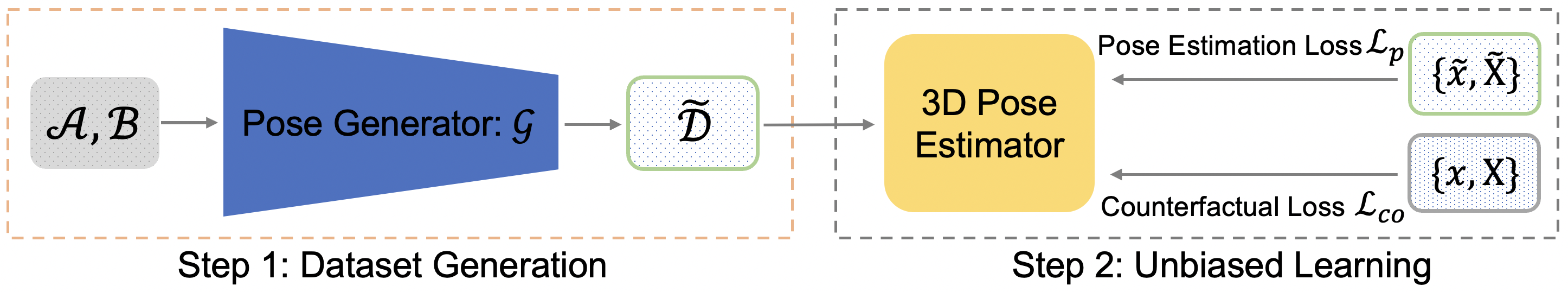}
    \caption{Two steps of \textit{PoseGU}, the first block is the training dataset generation and the second block shows the unbiased learning.}
    \label{fig:posegu}
\end{figure}

The overall training tasks of our \textit{PoseGU} is depicted in Fig~\ref{fig:posegu}.
We first use our proposed pose generator $\mathcal{G}$ to generate $K$ number of 3D pose coordinates as training dataset $\widetilde{\mathcal{D}}$ by giving a set of angle matrices $\mathcal{A} = \left\{\mathbf{A}_{k} \mid \forall k \in\{1, \cdots, K\}\right\}$ and a set of bone length vectors $\mathcal{B} = \left\{\boldsymbol{l}_{k} \mid \forall k \in\{1, \cdots, K\}\right\}$; 
Then we use generated 3D pose dataset $\widetilde{\mathcal{D}}$ to train a pose estimator $\mathcal{P}$ with pose estimation loss $\mathcal{L}_{\mathcal{P}}$ and counterfactual loss $\mathcal{L}_{co}$ for estimating 3D poses from 2D poses. 

\subsection{3D Human Pose Generator}
Our 3D pose generator $\mathcal{G}$ aims to generate diverse 3D human poses $\widetilde{\mathbf{X}}$ by giving rotation angle matrix $\mathbf{A}$ and bone length vector $\boldsymbol{l}$. 
To achieve this goal, we firstly introduce the forward kinematics  transformation~\cite{forwardkinematics} that generates 3D human poses by giving a random rotation angle matrix. 
Moreover, as random rotation angle matrices could introduce extremely hard cases, we further compute the joint rotation angle ranges from existing 3D poses to regulate input $\mathbf{A}$ in a strict range by taking advantage of inverse kinematics. 

\begin{figure}[h]
    \centering
    \includegraphics[width=1\linewidth]{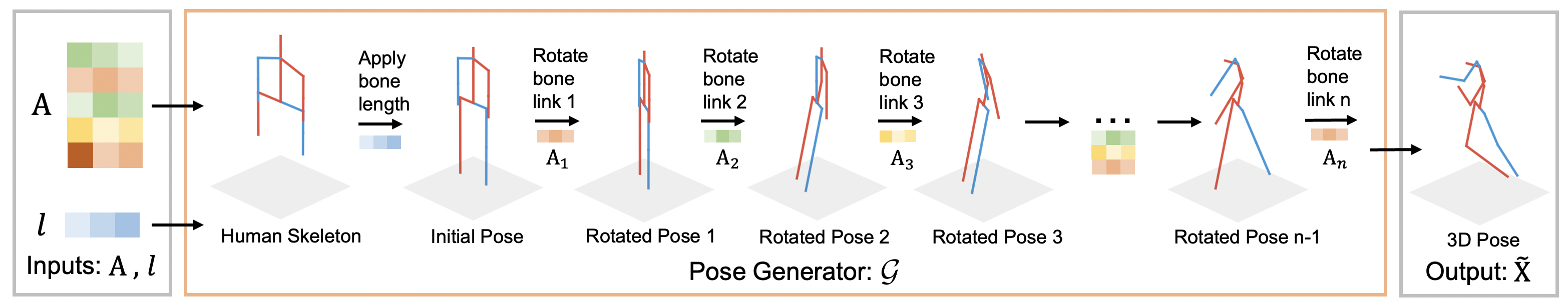}
    \caption{General pipeline of pose generator, our pose generator $\mathcal{G}$ takes rotation angle matrix $\mathbf{A}$ and bone length vector $\boldsymbol{l}$ as inputs. Human skeleton is a human shape with same unit bone length, and the initial pose is applied bone length vector to it. The output 3D pose $\widetilde{\mathbf{X}}$ is obtained by applying rotation transformation to bone links.}
    \label{fig:posegenerator}
\end{figure}

\subsubsection{Forward Kinematics for Generating 3D Human Pose}
Human pose has a kinematic configuration similar to industrial robot, we can utilize the Forward Kinematics equations from robotics~\cite{forwardkinematics} to compute the 3D position of each joint in camera coordinate system. 
However, due to no prismatic joint exits in the human body, we only consider the rotation transformation in this work. 
Here we define the joint position in camera coordinate system as $\boldsymbol{c}$, a bone vector $\boldsymbol{b}_{k}=\boldsymbol{c}_{r}-\boldsymbol{c}_{t}$ is defined as the vector between $r$-th and $t$-th joint. 
As we all know, the movement of human limbs follows a typical hierarchical skeletal movement, which means all child bones need to follow the parent bone moves, therefore, we need to apply the rotation matrix~\cite{rotationmatrix} on bone links to achieve hierarchical movement. 
As shown in Fig~\ref{fig:posegenerator}, for each bone link, we use $\mathbf{A}_{i}^{x}$, $\mathbf{A}_{i}^{y}$, $\mathbf{A}_{i}^{z}$ about $x$-$y$-$z$ axis to indicate joint rotation angles. 
Then we feed the rotation angles into the rotation transformation matrix $\mathbf{R}$ to convert angles into rotation matrix about $x$-$y$-$z$ axis respectively, and the complete rotation matrix is calculated by:

\begin{equation}
\mathbf{V}_{i}\left(\mathbf{A}_{i}\right)=\mathbf{R}_{x}\left(\mathbf{A}_{i}^{x}\right) \mathbf{R}_{y}\left(\mathbf{A}_{i}^{y}\right) \mathbf{R}_{z}\left(\mathbf{A}_{i}^{z}\right)
\end{equation}
where $\mathbf{V}_{i}$ denotes the rotation matrix for bone vector $\boldsymbol{b}_i$ and $\mathbf{R}_{k}$ denotes rotation transformation matrix about axis $k$. Then we can obtain the kinematics equations of a serial chain of $n$ bone links as follows:

\begin{equation}
\mathbf{T}_{n}=\mathbf{V}_{1} \mathbf{V}_{2} \ldots{ } \mathbf{V}_{n}
\end{equation}

The $\mathbf{T}_{n}$ denotes the transformation matrix on bone links with $n$ bone vectors and the rotated bone vector $\tilde{\boldsymbol{b}}$ can be obtained by:

\begin{equation}
\tilde{\boldsymbol{b}}_{i}=\mathbf{T}_{n} \boldsymbol{b}_{i}
\end{equation}

Then we can obtain the rotated bone matrix $\widetilde{\mathbf{B}} \in \mathbb{R}^{3 \times M}$ by concatenating $M$ bone vectors $\widetilde{\mathbf{B}}=\left(\tilde{\boldsymbol{b}}_{1}, \tilde{\boldsymbol{b}}_{2}, \cdots, \tilde{\boldsymbol{b}}_{M}\right)$, and the generated $\widetilde{\mathbf{X}}$ is converted by adding the root joint coordinate to $\widetilde{\mathbf{B}}$. 
We then introduce inverse kinematics that computes the joint rotation angle ranges from existing 3D poses to regulate input $\mathbf{A}$ in a strict range. 

\subsubsection{Inverse Kinematics for Computing Rotation Angle Matrix} 
Inverse kinematics is the reverse process of forward kinematics, which can compute the joint rotation angles from given poses. 
To ensure our $\mathcal{G}$ to generate poses have sufficient diversity but avoid generating implausible poses (i.e., pose not exits in reality) that hurt the training process, we need to regulate the random $\mathbf{A}$ to fall in a reasonable range. To obtain such range for regulating generated poses, we use inverse kinematics to deduce the joint angles from a given 3D pose. 
The overall pipeline of inverse kinematics is: we initially convert a given 3D pose into bone vectors, then, we project each bone vector on $x$-$y$-$z$ axis in its local coordinate system respectively and the corresponding rotation angle vector $\mathbf{A}_{i}$ can be computed by:

\begin{equation}
\mathbf{A}_{i}=\left\{\arccos \left(\frac{\boldsymbol{v}_{x}}{\boldsymbol{b}_{i}}\right), \arccos \left(\frac{\boldsymbol{v}_{y}}{\boldsymbol{b}_{i}}\right), \arccos \left(\frac{\boldsymbol{v}_{z}}{\boldsymbol{b}_{i}}\right)\right\}
\end{equation}
where the $\boldsymbol{v}_{k}$ is projection vector of $\boldsymbol{b}_{i}$ on axis $k$, and we can use the obtained rotation angles to regulate the rotation range for each bone. 
With our inverse kinematic based method, we can ensure that almost all generated 3D poses are reasonable and diverse.

\subsection{Construction Dataset Using 3D Pose Sequences} 
Generally, the 3D poses in ground truth dataset are represented as several daily activities (e.g., walking, phoning, sitting), which remain spatial-temporal information. 
Our experiments found that the training of estimators on spatial-temporal discontinuous pose data is unstable and would lead to worse performance on estimating 3D poses. 
Therefore, we aim to develop an architecture to generate sequences of 3D poses data as a training dataset rather than purely generate discontinuous 3D poses. 
Because our $\mathcal{G}$ only takes $\mathbf{A}$ and $\boldsymbol{l}$ as inputs, our goal can be simplified as generating a set of $\mathbf{A}$ that continuously changes over time and a set of corresponding $\boldsymbol{l}$. 
First we follow the action category to randomly sample few seed samples in the ground truth dataset. 
Then we use our proposed inverse kinematics method to extract their corresponding joint angle matrix, and a specific rotation angle range for each joint is made based on the extracted joint angle by different action categories. 
In the following steps, a random rotation angle matrix can be generated within this range. 

\begin{figure}[h]
    \centering
    \includegraphics[width=1\linewidth]{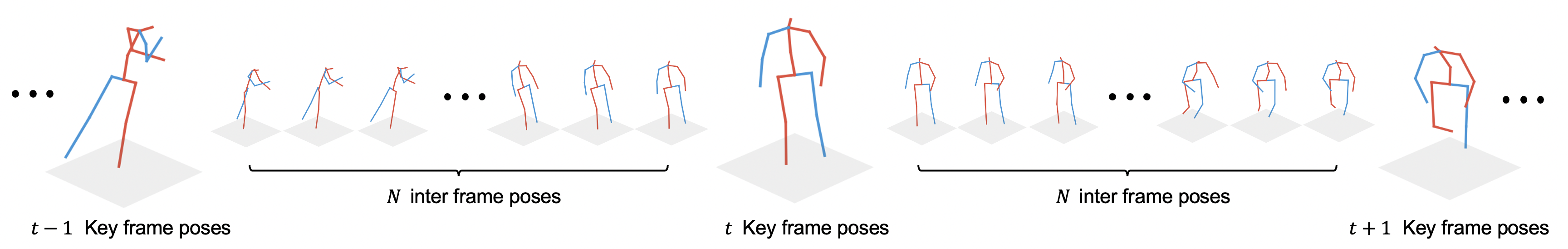}
    \caption{Visualize for generating 3D pose dataset, between each two adjacent key frame poses, we fulfill with $N$ inter frame poses by using linear interpolation method. These pose examples are generated by our pose generator.}
    \label{fig:posearcticture}
\end{figure}

Inspired by ``key frame''~\cite{keyframe} in video sequence, we first generate a set of $\left\{\mathbf{A}_{t} \mid \forall t \in\{1, \cdots, T\}\right\}$ with random values for generating $T$ key frame poses. 
Then we fulfill each two adjacent key frames with $N$ inter frames by using linear interpolation method, and the $n$-th $\mathbf{A}_n$ is given by:

\begin{equation}
\mathbf{A}_{n}=\mathbf{A}_{t-1}+n \times \frac{\mathbf{A}_{t}-\mathbf{A}_{t-1}}{N}
\end{equation}
where $\mathbf{A}_{t}$ and $\mathbf{A}_{t-1}$ are the $t$-th key frame and its previous adjacent key frame. 
With this method, the $\mathcal{A}$ which contains $K=N \cdot T$ rotation angle matrices can be obtained. 
To simulate poses with various camera views and body sizes, we also randomly add rotation angles to whole body, and we prepare few $\boldsymbol{l}$ templates that are extracted from seed samples to obtain $\mathcal{B}$ by random sampling. To this end, the $\widetilde{\mathcal{D}}$ can be generated by feeding $\mathcal{A}$ and $\mathcal{B}$ to pose generator $\mathcal{G}$.

\subsection{Counterfactual Risk Minimization for Unbiased Learning}
Our proposed 3D pose generator $\mathcal{G}$ can generate arbitrary 3D poses that may cover all possible poses in reality. Meanwhile, it ensures the generated dataset maintain rich pose diversity. 
However, we found that the pose estimators trained on $\widetilde{\mathcal{D}}$ (i.e., our generated dataset) cannot perform as well as on ground truth test dataset, at the same time, increasing the amount of training data only has a limit improvement. 
This means it is not an efficient way to improve the performance of a 3D pose estimator by only expanding the training dataset. 
Through the observation of the data distribution of joint positions (cf. Fig~\ref{fig:datadistribute}), we found that the distribution of $\widetilde{\mathcal{D}}$ (i.e., the generated dataset) differs from the distribution retained in ${\mathcal{D}}$ (i.e., the ground truth dataset). 
This is because poses in ${\mathcal{D}}$ display a skewed popularity distribution across different action categories, e.g., standing poses exist in most actions, but sitting poses only exist in site-related actions. 
Thus, the ${\mathcal{D}}$ suffers from a strong build-in ``popularity bias'' issue~\cite{Unbiasedtraining3}.
Since the $\widetilde{\mathcal{D}}$ from our pose generator is generated by taking seed samples from each action category as the input, the $\widetilde{\mathcal{D}}$ can also suffer from the trap of popularity bias.
In such case, training pose estimators directly on $\widetilde{\mathcal{D}}$ (i.e., biased) cannot reach an unbiased estimation, leading to sub-optimal performance of estimating 3D poses. 
Hence, we aim to develop an unbiased leaning objective for our pose estimator based on the pose distributions of the ground truth dataset ${\mathcal{D}}$ and the generated dataset $\widetilde{\mathcal{D}}$.
Specifically,  we firstly compute the propensity score of each input 2D pose $\tilde{\boldsymbol{x}}$ to reveal its identical data distribution. 
We then use these propensity scores to perform a Counterfactual Risk Minimization (CRM), such that the data distribution mismatch between ${\mathcal{D}}$ and $\widetilde{\mathcal{D}}$ can be removed. 

\subsubsection{Data distribution} 
We aim to develop a method to compute the probability of a pose appearing in the camera coordinate system, i.e., the propensity score. 
Since the joint positions are not continuous in the camera coordinate system, we resort to the histogram map to organize joint data to show their frequency distribution. 
Through our experiments, we found that 2D histogram map is sufficient to show the frequency distribution of joint data. Compared with 3D histogram map, 2D histogram map is more effective on dealing with small-scale ground truth dataset (i.e., 3DPW) and can save more computation resources. 
Here we define propensity score $\rho$ for an input 2D pose $\boldsymbol{x}$ in ${\mathcal{D}}$ as:

\begin{equation}
\rho(\boldsymbol{x})=\frac{1}{J} \sum_{i=1}^{J} H_{i}\left(x_{i}, y_{i}\right)
\end{equation}
where the $H_i$ is the 2D histogram operator~\cite{2Dhistrogram} of $i$-th joint in $\mathbf{X}$, it outputs the frequency of an input joint position $\left(x_{i}, y_{i}\right)$. In our work, we randomly take 25 \% samples on ${\mathcal{D}}$ to generate $H$. 
Analogously, the propensity score $\tilde{\rho}$ for $\boldsymbol{x}$ in the generated dataset ${\widetilde{\mathcal{D}}}$ is defined as:

\begin{equation}
\tilde{\rho}(\tilde{\boldsymbol{x}})=\frac{1}{J} \sum_{i=1}^{J} \widetilde{H}_{i}\left(\tilde{x}_{i}, \tilde{y}_{i}\right)
\end{equation}
where $\widetilde{H}_i$ denotes the 2D histogram operator of $i$-th joint in $\widetilde{\mathbf{X}}$, and it is generated with all samples in ${\widetilde{\mathcal{D}}}$. 

\subsubsection{Counterfactual Risk Minimization} 
An empirical study (cf. Figure~\ref{fig:datadistribute}) conducted on the data distributions of generated datasets (i.e., $\widetilde{\mathcal{D}}$) and ground truth datasets (i.e., ${\mathcal{D}}$) validates that the generated dataset holds a different distribution from the original data distribution in the ground truth dataset, which would pose severe bias issues to the downstream task (i.e., pose estimation).
Thus, directly optimizing the downstream pose estimator using the biased $\widetilde{\mathcal{D}}$ cannot reach an unbiased estimation, leading to sub-optimal performance of estimating 3D poses~\cite{crmcv1}.
To achieve the unbiased learning of pose estimator, we need to correct the discrepancy between the data distribution of ${\mathcal{D}}$ and the data distribution of $\widetilde{\mathcal{D}}$. 
Here we apply \textit{Counterfactual Risk Minimization} (CRM)~\cite{crmdebias1,CRMfairness} to include a counterfactual loss $\mathcal{L}_{co}$ computed on the propensity scores of ${\mathcal{D}}$ and $\widetilde{\mathcal{D}}$.
The CRM implements an Inverse Propensity Scoring (IPS) estimator to directly model the distribution shift in its objective.

Formally, we first design the IPS estimator based on propensity scores of ${\mathcal{D}}$ and $\widetilde{\mathcal{D}}$. 
However, the conventional IPS estimator suffers from a large variance due to the ``propensity overfitting'' issue~\cite{CRMfairness,CRMfairness2,Correctmismatch}, therefore,
we additionally include a clipped estimator that can restrict the propensity ratios (i.e., importance weight) to a maximum value~\cite{bottou2013counterfactual}. 
The core idea is to regulate large weights that are associated with rare poses in a training dataset. The clipped estimator (cIPS) can be represented as:

\begin{equation}\label{eq:cips}
L_{c I P S}\left(\tilde{\boldsymbol{x}_i}\right)=\frac{1}{N} \sum_{i=1}^{N} \min \left\{\frac{\tilde{\rho}\left(\tilde{\boldsymbol{x}}_{i}\right)}{\rho\left(\tilde{\boldsymbol{x}}_{i}\right)}, c\right\}
\end{equation}
where $c$ is a constant for regulating the importance weight $\frac{\tilde{\rho}\left(\tilde{x}_{i}\right)}{\rho\left(\tilde{x}_{i}\right)}$ to maximum at $c$, and $N$ indicates the number of input poses $\tilde{\boldsymbol{x}}$ in a training batch. 

\subsection{Training Loss}
\subsubsection{Pose Estimation Loss} We resort to the mean squared errors (MSE) as our pose estimation loss, which is used for estimating loss between estimated pose $\mathbf{X}^{\prime}$ and ground truth poses $\widetilde{\mathbf{X}}$ in $\widetilde{\mathcal{D}}$:

\begin{equation}
\label{LP}
\mathcal{L}_{\mathcal{P}}\left(\mathcal{P}_{\theta}, {\widetilde{\mathcal{X}}}\right)=\|\mathbf{X}^{\prime}-\widetilde{\mathbf{X}}\|_{2}^{2}
\end{equation}

We train the pose estimator $\mathcal{P}$ by only using the pose pair $\widetilde{\mathcal{X}}$ in our generated $\widetilde{\mathcal{D}}$, which can lead $\mathcal{P}$ have a competitive performance on ground truth test set. 

\subsubsection{Counterfactual Learning Loss} 
With the designed cIPS estimator, the CRM uses this cIPS estimator to re-weight the original estimation loss (cf. Eq.~\eqref{obj2}) according to the propensity score of input 2D pose $\tilde{\boldsymbol{x}}$ under $\widetilde{\mathcal{D}}$ and ${\mathcal{D}}$. 
Plugging Eq.~\eqref{eq:cips} into Eq.~\eqref{obj2}, the counterfactual learning loss can be written as:
\begin{equation}
\label{Lub}
\mathcal{L}_{co}=\sum_{i=1} ^ N L_{c I P S}\left(\tilde{\boldsymbol{x}_i}\right)\cdot\mathcal{L}_{\mathcal{P}}\left(\mathcal{P}, {\mathcal{X}}\right)
\end{equation}
where the $L_{c I P S}\left(\tilde{\boldsymbol{x}}\right)$ is the cIPS operator, $\tilde{\boldsymbol{x}}$ and ${\mathcal{X}}$ denote the input data on generated dataset $\widetilde{\mathcal{D}}$ and sampled pose pair on ground truth dataset $\mathcal{D}$. 

\subsubsection{End-to-End Training Strategy} With the model-agnostic pose training dataset $\widetilde{\mathcal{D}}$, the $\mathcal{P}$ can be trained end-to-end by using any downstream 3D pose estimation methods. We update its parameter by minimizing the overall loss function:

\begin{equation}
\mathcal{L}_{A} = \min _{\theta}\left( \mathcal{L}_{\mathcal{P}}\left(\mathcal{P}_{\theta}, \widetilde{\mathcal{X}}\right) + \mathcal{L}_{co}\left(\mathcal{P}, \tilde{\boldsymbol{x}},\mathcal{X}\right)\right)
\end{equation}
where $\mathcal{L}_{A}$ is the overall loss function by combining Eq.~\eqref{LP}) and Eq.~\eqref{Lub}).

\section{Experiments}
Without specific mention, we first randomly take 20 seed samples by each action category throughout our experiments. 
Based on these seed samples, we generate the corresponding 3D pose training dataset by using our proposed pose generator. 
Then we randomly sample 25\% training data from ground truth source dataset to implement our unbiased learning method to train an estimator, and use ground truth joints in test set for evaluation. 

\subsection{Datasets} 
To fully evaluate our \textit{PoseGU}, we implement our experiments on three popular benchmark datasets for quantitative evaluation: Human3.6M~\cite{H36M}, MPI-INF-3DHP~\cite{3DHP} and 3DPW~\cite{3DPW}, and two datasets for unseen poses visualization: MPII~\cite{mpii} and LSP~\cite{lsp}. 
\begin{itemize}
    \item \textbf{Human3.6M}~\cite{H36M}: is the largest 3D human pose estimation dataset which uses 3.6 million images to represent 15 categories of daily activities (e.g., Sitting, Walking and Phoning). 
    We evaluate our \textit{PoseGU} on Human3.6M using two evaluation metrics: Mean Per Joint Position Error (MPJPE) in millimeter, and MPJPE under rigid alignment transformation (P-MPJPE).
    \item \textbf{MPI-INF-3DHP (3DHP)}~\cite{3DHP}: has 1.3 million frames for 3D pose estimation. It has more diverse poses compared with Human3.6M, we use its test set to evaluate the generalization ability of the trained estimators. We test \textit{PoseGU} on the 3DHP using three evaluation metrics: MPJPE, Percentage of Correctly Positioned Key points (PCK) and Area Under Curve (AUC).
    \item \textbf{3DPW}~\cite{3DPW}: is an in-the-wild dataset that contains complicated poses and scenes. To evaluate the generalization ability of \textit{PoseGU} to challenging in-the-wild scenarios, we use its test set for evaluation by using P-MPJPE.
    \item \textbf{MPII}~\cite{mpii} and \textbf{LSP}~\cite{lsp}: are two widely used in-the-wild datasets with only 2D key joint annotations. 
    We extract unseen poses in these two datasets to qualitatively evaluate the model generalization.  
\end{itemize}

\subsection{Results} 
We use the generated dataset based on seed samples extracted from Human3.6M as the training dataset. 
We resort to the single-frame version of ST-GCN~\cite{stgcn} as the 3D pose estimator, and train it on the training dataset (i.e., the generated dataset) to test the performance of our \textit{PoseGU}. 

\begin{table}[h]
	\small
	\centering
	\caption{\textbf{Results on Human3.6M} in terms of average MPJPE. HR denotes 2D poses from HR-Net as inputs and GT denotes the ground truth 2D key joints. Best results are shown in \textbf{bold}.}
	\resizebox{95mm}{!}{
	\begin{tabular}{l|cc}
		\specialrule{1pt}{1pt}{1pt}
		Method &  MPJPE-HR~($\downarrow$) & MPJPE-GT~($\downarrow$)  \\
		\hline
		\rowcolor{grayLightDark}
        Yang et al. (CVPR’18)  ~\cite{adver2}   & 58.6 & -  \\
		\rowcolor{grayLight}
		Sharma et al. (CVPR’19) ~\cite{stateofart-sharma}   & 58.0 & -  \\
		\rowcolor{grayLightDark}
		Zhao et al. (CVPR’19) ~\cite{semigcn}  & 57.6 & 43.8   \\
		\rowcolor{grayLight}
		Moon et al. (ICCV’19) ~\cite{stateofart-moon2019camera} & 54.4 & 35.2 \\
		\rowcolor{grayLightDark}
		Li et al. (CVPR’20) ~\cite{aug1} & 50.9 & \textbf{34.5} \\
		\rowcolor{grayLight}
		Gong et al. (CVPR’21) ~\cite{poseaug}  & 50.2 & 39.1  \\
		\hline
		\rowcolor{grayLightDark}
		PoseGU    & \textbf{49.6} & 37.4   \\
		\hline
		\specialrule{1pt}{1pt}{2pt}
	\end{tabular}}
	\label{tab:comparewithH36m}
\end{table} 

\subsubsection{Results on Human3.6M} 
We compare our trained pose estimator with state-of-the-art (SOTA) methods~\cite{adver2,stateofart-sharma,semigcn,stateofart-moon2019camera,aug1,poseaug} on Human3.6M. 
In this experiment, we use pre-processed 2D poses from HR-Net~\cite{HR-Net} and ground truth 2D key joints as inputs, respectively. 
As shown in Table~\ref{tab:comparewithH36m}, our method outperforms SOTA methods in terms of HR-Net. 
Although our method achieves higher MPJPE with ground truth 2D pose inputs than the methods proposed in~\cite{aug1,stateofart-moon2019camera}, we only use 25\% training data in Human3.6M for unbiased learning. 
This clearly verifies the advantage of our \textit{PoseGU} - it only requires a small amount of ground truth data to achieve optimal performance on 3D pose estimation. 

\subsubsection{Results on 3DHP (cross-scenario)} 
We then evaluate our model generalization on cross-scenario dataset by comparing \textit{PoseGU} with various SOTA methods, including the latest offline pose augmentation method~\cite{aug1}, elaborated designed networks~\cite{LCN,SRNet}, weakly-supervised learning approaches~\cite{HMR,adver1,adver2} and directly trained on 3DHP~\cite{LCR-Net,Multiperson,VNect,OriNet}. 
It can be observed from Table~\ref{tab:3DHPcorssscenarioresult}, our \textit{PoseGU} outperforms all SOTA methods by a large margin in terms of all evaluation metrics. 
This strongly verifies our \textit{PoseGU} is able to generate more diverse 3D poses than Human3.6M and achieve a better generalization ability.

\begin{table}[h]
	\small
	\centering
	\caption{\textbf{Results on 3DHP} in terms of PCK, AUC and MPJPE. CE indicates cross-scenario evaluation. Best results are shown in \textbf{bold}.}
	\resizebox{80mm}{!}{
	\begin{tabular}{l|c|ccc}
		\specialrule{1pt}{1pt}{1pt}
		Method & CE & PCK~($\uparrow$) & AUC~($\uparrow$) & MPJPE~($\downarrow$) \\
		\hline
		\rowcolor{grayLightDark}
        LCR-Net~\cite{LCR-Net} & & 59.6 & 27.6 & 158.4 \\
        \rowcolor{grayLight}
        Multi Person ~\cite{Multiperson} & & 75.2 & 37.8 & 122.2 \\
        \rowcolor{grayLightDark}
        VNect~\cite{VNect}  & & 76.6 & 40.4 & 124.7 \\
        \rowcolor{grayLight}
        OriNet~\cite{OriNet}  & & 81.8 & 45.2 & 89.4 \\
        \hline
        \rowcolor{grayLightDark}
        Yang et al.~\cite{adver2} & \checkmark & 69 & 32 & - \\ 
        \rowcolor{grayLight}
        LCN~\cite{LCN}  & \checkmark & 74 & 36.7 & - \\    
        \rowcolor{grayLightDark}
        HMR~\cite{HMR}  & \checkmark & 77.1 & 40.7 & 113.2 \\  
        \rowcolor{grayLight}
        SRNet~\cite{SRNet}  & \checkmark & 77.6 & 43.8 & - \\
        \rowcolor{grayLightDark}
        Li et al.~\cite{aug1} & \checkmark & 81.2 & 46.1 & 99.7 \\
        \rowcolor{grayLight}
        RepNet~\cite{adver1} & \checkmark & 81.8 & 54.8 & 92.5 \\
		\hline
		\rowcolor{grayLightDark}
		PoseGU & \checkmark  & \textbf{86.3} & \textbf{55.1}  & \textbf{79.1} \\
		\hline
		\specialrule{1pt}{1pt}{2pt}	
	\end{tabular}}
	\label{tab:3DHPcorssscenarioresult}
\end{table} 

\subsubsection{Qualitative Results} 
\begin{figure}[h]
    \centering
    \includegraphics[width=1\linewidth]{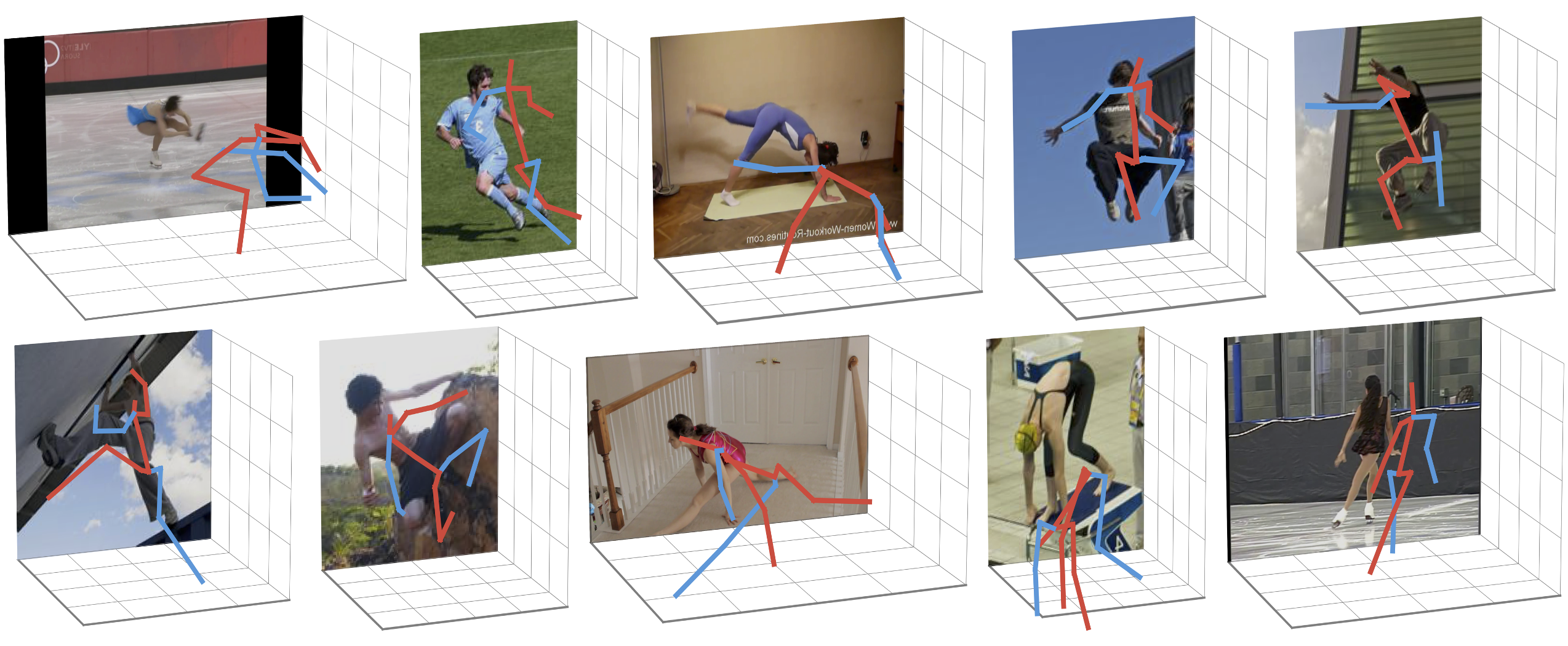}
    \caption{Qualitative results of estimated 3D poses from unseen poses in MPII and LSP, the 3D pose estimator is trained by our \textit{PoseGU}.}
    \label{fig:quantitiveresult}
\end{figure}

For qualitative evaluation through visualizing the estimated 3D poses from unseen poses, we use two challenging datasets: MPII and LSP, which have large varieties regarding camera view, body size, and postures. 
As shown in Fig~\ref{fig:quantitiveresult}, we can observe that our \textit{PoseGU} performs well on unseen poses, even for complicated poses. 

\subsection{Analysis of PoseGU}
We conduct an in-depth analysis of the effectiveness of our \textit{PoseGU} on improving model flexibility, mitigating bias issues, enhancing model generalization and model stability. 
We test the performance of \textit{PoseGU} in three scenarios which have ground truth 3D pose datasets by using four 3D pose estimators: 1) MLP~\cite{mlp}, 2) VideoPose~\cite{videopose}, 3) SemGCN~\cite{semigcn} (1-frame), and 4) ST-GCN~\cite{stgcn} (1-frame) and show our observations in this section. 

\begin{table}[h]
	\small
	\centering
	\caption{Analysis on the model flexibility to Human3.6M, 3DHP and 3DPW scenarios. GT, PoseGU indicate estimators trained on ground truth dataset, and \textit{PoseGU} respectively. Best results are shown in \textbf{bold}.}
	\resizebox{100mm}{!}{
	\begin{tabular}{l|c|c|cc|ccc|c}
		\specialrule{1pt}{1pt}{1pt}
		\multicolumn{3}{c|}{} & \multicolumn{2}{c|}{Human3.6M}  & \multicolumn{3}{c|}{3DHP} & \multicolumn{1}{c}{3DPW}\\
		\hline 
        Method & \multicolumn{1}{c|}{GT} & \multicolumn{1}{c|}{PoseGU} & \multicolumn{1}{c}{MPJPE} & \multicolumn{1}{c|}{P-MPJPE} & \multicolumn{1}{c}{PCK} & \multicolumn{1}{c}{AUC} &  \multicolumn{1}{c|}{MPJPE} &  \multicolumn{1}{c}{P-MPJPE}\\ 
		\hline
		MLP & \checkmark &    & 42.7 & 32.9 & 89.9 & 60.5 & 67.2 & 43.0 \\	
		\rowcolor{grayLightDark}
		\cite{mlp} & & \checkmark & \textbf{39.6} & \textbf{32.5} & \textbf{91.1} & \textbf{63.3}  & \textbf{59.2} & \textbf{36.2} \\
		\hline
		SemGCN & \checkmark &    & 44.6 & 33.3 & 89.1 & 55.8 & 75.2 & 38.9\\	
		\rowcolor{grayLightDark}
		\cite{semigcn} &  & \checkmark  & \textbf{39.2} & \textbf{32.2}  & \textbf{89.6} & \textbf{60.9} & \textbf{66.5}  & \textbf{35.1} \\
		\hline
		VPose& \checkmark &  & 41.8 & \textbf{32.8} & 89.6 & 60.8 & 67.3 & 42.7 \\	
		\rowcolor{grayLightDark}
		\cite{videopose}(1-frame) &  & \checkmark & \textbf{40.6} & 33.4  & \textbf{90.8} & \textbf{65.0} & \textbf{59.0}  & \textbf{37.8} \\
		\hline
		ST-GCN & \checkmark &  & 40.7 & 32.5 & 88.6 & 57.8 & 73.2 & 39.2\\	
		\rowcolor{grayLightDark}
		\cite{stgcn}(1-frame)  &  & \checkmark & \textbf{37.4} & \textbf{32.1}  & \textbf{89.9} & \textbf{60.2}  & \textbf{67.0}  & \textbf{35.0}  \\
		\hline
		\specialrule{1pt}{1pt}{2pt}	
	\end{tabular}}
	\label{tab:intrasenario}
\end{table}

\subsubsection{Model Flexibility}
To evaluate the flexibility of our \textit{PoseGU} in applying in different scenarios and pose estimators.
We train our \textit{PoseGU} with four different estimators, MLP~\cite{mlp}, VideoPose~\cite{videopose}, SemGCN~\cite{semigcn} and ST-GCN~\cite{stgcn} with three benchmark datasets (i.e., Human3.6M, 3DHP, 3DPW) and show the comparison in Table~\ref{tab:intrasenario}. 
Analyzing Table~\ref{tab:intrasenario}, we observe that all estimators trained with our \textit{PoseGU} achieve an optimal performance than their counterparts trained on ground truth datasets.
As for VPose in terms of P-MPJPE, when we use more data in ground truth dataset for unbiased learning (say 50\%), our model achieves 39.8 MPJPE and 32.5 P-MPJPE, respectively.
For the sparse dataset 3DPW, we generate 200K 3D poses as the training dataset. 
From the results in Table~\ref{tab:intrasenario}, we can observe that our generated training dataset with unbiased learning approach can achieve much lower values on P-MPJPE of all estimators by a large margin. 
This means our \textit{PoseGU} is applicable to various scenarios and pose estimators.

\subsubsection{Bias Alleviation} 
To validate the effectiveness of our unbiased learning approach on alleviating the bias issue, we implement experiments to compare the performance of estimators trained on our generated dataset with and without our unbiased learning approach. 
As shown in Table~\ref{tab:analysisunbiased}, we can observe that all estimators trained on our generated dataset have competitive performance.
When we apply unbiased learning to our generated dataset, it brings significant improvements on estimating 3D poses in all scenarios. This clearly verifies our unbiased learning approach improves the performance of all estimators thanks to its ability of alleviating the bias issue caused by the distribution mismatch. 

\begin{table}[h]
	\small
	\centering
	\caption{Effective of unbiased learning on \textit{PoseGU}. PG, PoseGU indicate the estimators trained on our generated dataset, and \textit{PoseGU} respectively. Best results in \textbf{bold}.}
	\resizebox{100mm}{!}{
	\begin{tabular}{l|c|c|cc|ccc|c}
		\specialrule{1pt}{1pt}{1pt}
		\multicolumn{3}{c|}{} & \multicolumn{2}{c|}{Human3.6M}  & \multicolumn{3}{c|}{3DHP} & \multicolumn{1}{c}{3DPW}\\
		\hline 
        Method  & \multicolumn{1}{c|}{PG} & \multicolumn{1}{c|}{PoseGU} & \multicolumn{1}{c}{MPJPE} & \multicolumn{1}{c|}{P-MPJPE} & \multicolumn{1}{c}{PCK} & \multicolumn{1}{c}{AUC} &  \multicolumn{1}{c|}{MPJPE} &  \multicolumn{1}{c}{P-MPJPE}\\ 
		\hline
		MLP & \checkmark &   & 59.5 & 44.5 & 87.2 & 60.2 & 70.5 & 50.9 \\	
		\rowcolor{grayLightDark}
		\cite{mlp}&  & \checkmark & \textbf{39.6} & \textbf{32.5} & \textbf{91.1} & \textbf{63.3}  & \textbf{59.2} & \textbf{36.2} \\
		\hline
		SemGCN & \checkmark &  & 59.8 & 40.7 & 88.1 & 55.2 & 76.5 & 52.1\\	
		\rowcolor{grayLightDark}
		\cite{semigcn}&  & \checkmark & \textbf{39.2} & \textbf{32.2}  & \textbf{89.6} & \textbf{60.9} & \textbf{66.5} & \textbf{35.1} \\
		\hline
		VPose & \checkmark &   & 60.6 & 44.8 & 87.4 & 60.6 & 69.9 & 52.7 \\	
		\rowcolor{grayLightDark}
		\cite{videopose}(1-frame)&  & \checkmark & \textbf{40.6} & \textbf{33.4}  & \textbf{90.8} & \textbf{65.0} & \textbf{59.0} & \textbf{37.8} \\
		\hline
		ST-GCN & \checkmark &  & 58.2 & 43.5 & 88.6 & 59.5
& 69.6 & 48.9\\	
		\rowcolor{grayLightDark}
		\cite{stgcn}(1-frame)&  & \checkmark & \textbf{37.4} & \textbf{32.1}  & \textbf{89.9} & \textbf{60.2}  & \textbf{67.0} & \textbf{35.0} \\
		\hline
		\specialrule{1pt}{1pt}{2pt}	
	\end{tabular}}
	\label{tab:analysisunbiased}
\end{table}

\subsubsection{Model Generalization} 
To validate our model generalization ability, we compare four estimators trained on Human3.6M (denoted as GT in Table~\ref{tab:analysiscross}), our generated dataset but without unbiased learning (denoted as PG) and PoseGU respectively, we evaluate these estimators on test sets of 3DHP and 3DPW.
As shown in Table~\ref{tab:analysiscross}, we can observe that the estimators trained on our generated dataset with unbiased learning approach achieve the best performance on all scenarios. 
Compared with the ones trained on Human3.6m directly, although our generated dataset leads to a sub-optimal prediction performance on Human3.6M test set due to the data distribution mismatch between Human3.6M and our generated dataset, our generated dataset leads to a better model generalization on cross-scenario. 
This clearly verifies our pose generator can generate sufficiently diverse and reasonable 3D pose data to train a pose estimator. Moreover, our unbiased learning method can further improve model generalization ability.

\begin{table}[h]
	\small
	\centering
	\caption{Model generalization evaluation on 3DHP and 3DPW, CE denotes cross-scenario. GT, PG, PoseGU indicate estimators trained on ground truth dataset, generated dataset, and \textit{PoseGU} respectively. Best results are shown in \textbf{bold}.}
	\resizebox{115mm}{!}{
	\begin{tabular}{l|c|c|c|cc|ccc|c}
		\specialrule{1pt}{1pt}{1pt}
		\multicolumn{4}{c|}{} & \multicolumn{2}{c|}{Human3.6M} & \multicolumn{3}{c|}{3DHP (CE)} & \multicolumn{1}{c}{3DPW (CE)} \\
		\hline 
        Method & \multicolumn{1}{c|}{GT} & \multicolumn{1}{c|}{PG} & \multicolumn{1}{c|}{PoseGU} & \multicolumn{1}{c}{MPJPE} & \multicolumn{1}{c|}{P-MPJPE} & \multicolumn{1}{c}{PCK} & \multicolumn{1}{c}{AUC} &  \multicolumn{1}{c|}{MPJPE} & \multicolumn{1}{c}{P-MPJPE}\\ 
		\hline
		 & \checkmark &  &  & 42.7 & 32.9 & 79.96 & 48.1 & 93.3 & 90.9\\	
		\rowcolor{grayLight}
		MLP~\cite{mlp} &  & \checkmark & & 59.5 & 44.5 & 85.1 & 55.0	& 81.7 & 90.3\\
		\rowcolor{grayLightDark}
		 &  &  & \checkmark & \textbf{39.6} & \textbf{32.5} & \textbf{85.6} & \textbf{56.4} & \textbf{76.8} & \textbf{87.0}\\
		\hline
		 & \checkmark &  & & 44.6 & 33.3 & 80.2 & 49.6	& 88.4 & 99.3\\	
		\rowcolor{grayLight}
		SemGCN~\cite{semigcn} &  & \checkmark &  & 59.8 & 40.7 & 84.3	& 52.8	& 87.0 & 99.0\\
		\rowcolor{grayLightDark}
	    & &  & \checkmark & \textbf{39.2} & \textbf{32.2} & \textbf{86.3} & \textbf{55.9} & \textbf{77.0} & \textbf{76.6}\\
		\hline
		 & \checkmark &  & & 41.8 & \textbf{32.8} & 81.6 & 48.7	& 92.4 & 95.0\\	
		\rowcolor{grayLight}
		VPose~\cite{videopose}(1-frame) & & \checkmark & & 60.6 & 44.8 & 85.7  &	55.7 & 80.3 & 92.5\\
		\rowcolor{grayLightDark}
	    &  &  & \checkmark & \textbf{40.6} & 33.4 & \textbf{86.3} & \textbf{57.2} & \textbf{75.0} & \textbf{92.3}\\
		\hline
		& \checkmark &  & & 40.7 & 32.5 & 83.3 & 51.5 & 83.1 & 97.1\\	
		\rowcolor{grayLight}
		ST-GCN~\cite{stgcn}(1-frame) &  & \checkmark & & 58.2 & 43.5 & 84.5 & 55.1 & 79.8 & 97.0\\
		\rowcolor{grayLightDark}
		&  &  & \checkmark & \textbf{37.4} & \textbf{32.1} & \textbf{86.3}	& \textbf{55.8}	& \textbf{79.1} & \textbf{84.2}\\
		\hline
		\specialrule{1pt}{1pt}{2pt}	
	\end{tabular}}
	\label{tab:analysiscross}
\end{table}

\begin{figure}[h]
    \centering
    \includegraphics[width=1\linewidth]{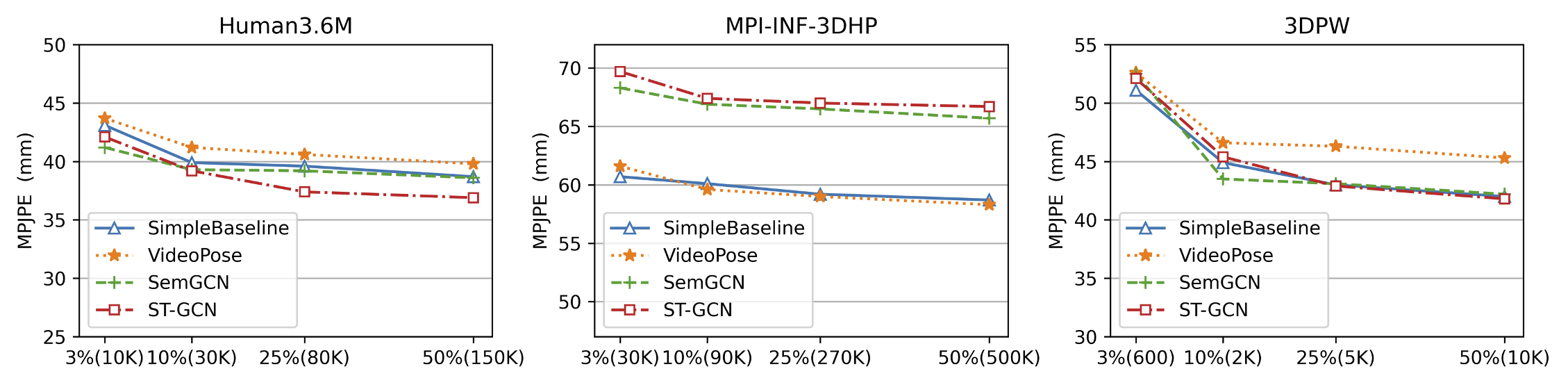}
    \caption{Model stability analysis of our \textit{PoseGU} on Human3.6M, 3DHP, and 3DPW. The vertical axis shows the MPJPE in millimeter and horizontal axis indicates the percentage of ground truth data usage and corresponding approximate 3D poses number.}
    \label{fig:ablationstudyCRM}
\end{figure}

\subsubsection{Model Stability} 
Here we sample 3\%, 10\%, 25\%, and 50\% of ground truth training data to implement unbiased learning on three benchmark datasets, and we evaluate these estimators on their corresponding test set in terms of MPJPE. 
As shown in Fig ~\ref{fig:ablationstudyCRM}, we can observe that the estimators train by our proposed unbiased learning method only requires very small amount of ground truth training data (less than 10\%) to achieve similar or even better performance than the ones trained on ground truth training data directly. In addition, the MPJPE of estimators decrease rapidly when we utilize more ground truth training data (less than 25\%). 
However, the performance of estimators only have a subtle improvement when we further increase the usage of ground truth training data (more than 25\%). This is because too little data makes it difficult to represent the distribution of ground truth dataset, the performance of estimator is close to best when there is enough data to represent the dataset distribution. 

\subsubsection{Effectiveness of Counterfactual Learning Loss} 
To evaluate the effectiveness of counterfactual learning loss, we set the cIPS operator in Eq.~\ref{Lub} as 1, the other settings remain the same as \textit{PoseGU}. 

\begin{table}[h]
	\small
	\centering
	\caption{Effective of counterfactual learning loss on \textit{PoseGU}. CL indicates the estimators trained within and without counterfactual learning loss respectively. Best results are shown in \textbf{bold}.}
	\resizebox{100mm}{!}{
	\begin{tabular}{l|c|cc|ccc|c}
		\specialrule{1pt}{1pt}{1pt}
		\multicolumn{2}{c|}{} & \multicolumn{2}{c|}{Human3.6M}  & \multicolumn{3}{c|}{3DHP} & \multicolumn{1}{c}{3DPW}\\
		\hline 
        Method  & \multicolumn{1}{c|}{CRM} & \multicolumn{1}{c}{MPJPE} & \multicolumn{1}{c|}{P-MPJPE} & \multicolumn{1}{c}{PCK} & \multicolumn{1}{c}{AUC} &  \multicolumn{1}{c|}{MPJPE} &  \multicolumn{1}{c}{P-MPJPE}\\ 
		\hline
		MLP & \checkmark  & \textbf{39.6} & \textbf{32.5} & \textbf{91.1} & \textbf{63.3} & \textbf{59.2} & \textbf{36.2} \\	
		\rowcolor{grayLightDark}
		\cite{mlp}&   & 43.7 & 35.6 & 89.9 & 61.6  & 64.9 & 39.9 \\
		\hline
		SemGCN & \checkmark  & \textbf{39.2} & \textbf{32.2} & \textbf{89.6} & \textbf{60.9} & \textbf{66.5} & \textbf{35.1}\\	
		\rowcolor{grayLightDark}
		\cite{semigcn}&  & 45.5 & 36.0  & 87.9 & 57.8 & 72.3 & 39.2 \\
		\hline
		VPose & \checkmark &  \textbf{40.6} & \textbf{33.4} & \textbf{90.8} & \textbf{65.0} & \textbf{59.0} & \textbf{37.8} \\	
		\rowcolor{grayLightDark}
		\cite{videopose}(1-frame)&  & 46.2 & 42.0  & 88.6 & 61.8 & 64.0 & 43.7 \\
		\hline
		ST-GCN & \checkmark & \textbf{37.4} & \textbf{32.1}  & \textbf{89.9} & \textbf{60.2} & \textbf{67.0} & \textbf{35.0}\\	
		\rowcolor{grayLightDark}
		\cite{stgcn}(1-frame)&  & 44.3 & 37.4  & 89.1 & 59.0  & 69.9 & 40.8 \\
		\hline
		\specialrule{1pt}{1pt}{2pt}	
	\end{tabular}}
	\label{tab:analysiscrm}
\end{table}

As shown in Table~\ref{tab:analysiscrm}, we can observe that our counterfactual learning loss can effectively improve the performance of pose estimators in terms of all evaluation protocols, especially in terms of MPJPE and P-MPJPE.
Compared with the PG column in Table~\ref{tab:analysisunbiased}, the estimators train without counterfactual learning loss also have a significant improvement, but cannot achieve the same level as \textit{PoseGU}. 
This means our counterfactual learning loss is able to re-weight the estimation loss on ground truth training data, and lead a optimal performance on estimating 3D poses and improve model generalization. 

\section{Conclusion}
In this paper, we develop a novel approach \textit{PoseGU} for 3D human pose estimation by using a model-agnostic training dataset generated by our novel 3D pose generator with an unbiased learning objective.
Our \textit{PoseGU} only requires minimum ground truth samples to generate a good enough training dataset to train a 3D pose estimator. 
Conservatively speaking, using less than 25\% of the full ground truth 3D pose dataset, our \textit{PoseGU} can achieve as good as or even better performance than using the full ground truth training set. 
The superior performance of this new 3D estimation approach is contributed largely by the higher level of diversity of our generated training dataset enabled by the novel idea used in the generator design, while our unbiased learning approach achieves an unbiased learning of estimator to guide a better performance of 3D pose estimation and better model generalization ability.



\clearpage
%
%
\bibliographystyle{splncs04}
\bibliography{egbib}
\end{document}